%% file: main.tex
\documentclass[10pt,twocolumn,letterpaper]{article}
\usepackage[T1]{fontenc}
\usepackage[pagenumbers]{cvpr}
\input{preamble}
\definecolor{cvprblue}{rgb}{0.21,0.49,0.74}
\usepackage[pagebackref,breaklinks,colorlinks,allcolors=cvprblue]{hyperref}
\title{FaceJev: Direct Visual Decisions for Face Forgery Detection}
\author{
Siran Peng\textsuperscript{1,2,$*$}\quad
Tianshuo Zhang\textsuperscript{2,1,$*$}\quad
Yibo Xue\textsuperscript{1,2}\quad
Ajian Liu\textsuperscript{1,2}\quad
Haoyuan Zhang\textsuperscript{2,1}\\
Weisong Zhao\textsuperscript{3}\quad
Shuhuan Chen\textsuperscript{3}\quad
Xiangyu Zhu\textsuperscript{1,2}\quad
Zhen Lei\textsuperscript{1,2,4,$\dagger$}\\
\textsuperscript{1}MAIS, CASIA\quad
\textsuperscript{2}SAI, UCAS\quad
\textsuperscript{3}IIE, CAS\quad
\textsuperscript{4}SCSE, FIE, M.U.S.T.\\
{\ttfamily\small\{pengsiran2023, zhangtianshuo2022, zhen.lei\}@ia.ac.cn}\\
{\small\textsuperscript{$*$}Equal contribution.\quad\textsuperscript{$\dagger$}Corresponding author.}
}
\date{}
\begin{document}
\maketitle
\input{sec/0_abstract}
\begin{figure}[t]
 \centering
 \includegraphics[width=0.90\columnwidth]{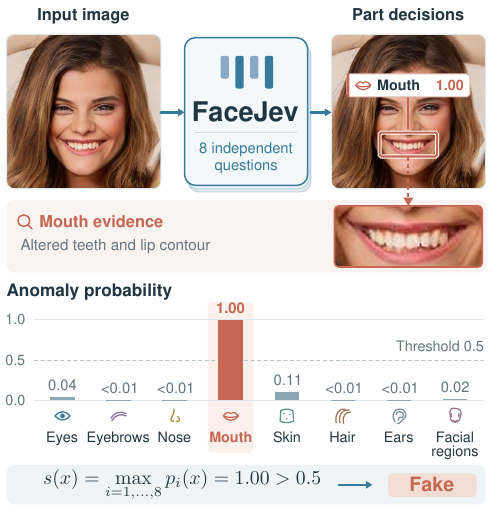}
 \caption{\textbf{FaceJev detects forgeries through direct visual decisions.} Eight independent questions share an image encoding, with anomaly probabilities read from native answer logits. An image is classified as forged if the maximum part probability reaches 0.5.}
 \label{fig:teaser}
\end{figure}
\input{sec/1_intro}
\input{figures/method_float}
\input{sec/2_related}
\input{sec/3_method}
\input{sec/4_dataset}
\input{sec/5_experiments}
\input{sec/6_conclusion}
{\small

\input{main.bbl}
}
\end{document}

%% file: preamble.tex
\usepackage{amsmath,amssymb,booktabs,multirow,array,tabularx}
\usepackage{graphicx}
\usepackage{colortbl}
\usepackage{microtype}
\usepackage{enumitem}
\definecolor{faceblue}{HTML}{27647C}
\definecolor{palerow}{HTML}{EAF2F6}
\definecolor{groupgray}{HTML}{F3F3F3}
\definecolor{deltagray}{HTML}{5F6B73}
\newcommand{\ind}{\mathbf{1}}

\newcommand{\best}[1]{\textbf{#1}}
\newcommand{\second}[1]{\underline{#1}}
\newcommand{\tablefont}{\small}
\newcolumntype{Y}{>{\centering\arraybackslash}X}
\newcommand{\grouprow}[2]{\rowcolor{groupgray}\multicolumn{#1}{l}{\textit{#2}}\\[1pt]}
\newcommand{\dn}[1]{\textcolor{deltagray}{$-$#1}}
\newcommand{\up}[1]{\textcolor{deltagray}{$+$#1}}
\DeclareMathOperator{\OR}{OR}

\setlist[itemize]{leftmargin=1.1em,topsep=2pt,itemsep=1pt,parsep=0pt}
\makeatletter
\renewcommand\paragraph{\@startsection{paragraph}{4}{\z@}{5pt plus 1pt minus 1pt}{-1em}{\normalfont\normalsize\bfseries}}
\makeatother

%% file: sec/0_abstract.tex
\begin{abstract}
Face forgery detection methods often perform well on seen manipulations but generalize poorly to unseen ones. Multimodal Large Language Models (MLLMs) can recognize facial parts and assess their visual plausibility. Existing MLLM-based methods typically express these judgments through textual explanations, but seldom treat individual part decisions as explicit learning targets. We present FaceJev, an MLLM-based method for face forgery detection through direct visual decisions. Following the Jev formulation, FaceJev answers eight independent questions about facial parts using a shared image encoding and reads anomaly probabilities from the native answer logits. A face is classified as forged if any part is predicted to be abnormal, directly linking the detection result to the part decisions. We construct PartForensics, a dataset of approximately 160k images with binary anomaly targets for the eight facial parts, to supervise these decisions. We further propose FS-GRPO, which optimizes each part decision through an exact reward comparison with its alternative, using both detection and part rewards. FaceJev outperforms the strongest baseline by 7.25 percentage points on unseen manipulations. Extensive ablation studies validate our design choices and the effectiveness of the main components.
\end{abstract}

%% file: sec/1_intro.tex
\section{Introduction}
\label{sec:intro}
Face forgery detection helps safeguard information integrity and public trust by identifying manipulated facial content. Existing methods can achieve high accuracy on seen manipulations, yet their performance often drops substantially on unseen ones~\cite{Shiohara_2022_CVPR,yermakov2025gend}. New methods for face swapping, reenactment, talking face generation, and diffusion editing can produce artifacts that differ from those represented in training data. Reliable detection therefore requires generalization to manipulation methods unseen in the training data.

Multimodal Large Language Models (MLLMs) can recognize facial parts and assess their visual plausibility, including abnormalities such as fused teeth, mismatched irises, or blurred lip boundaries. Existing MLLM-based methods incorporate these capabilities into detection through textual explanations~\cite{wen2025fakevlm,peng2025mllm} or forensic reasoning~\cite{tan2025veritas,li2026vigil}. Training on textual explanations supervises both the visual judgments and the wording used to express them, although generating that wording is not required to determine whether a facial part is abnormal. Individual part decisions are seldom treated as explicit learning targets that directly determine the detection result for the input image.

To learn these part decisions directly, we introduce \textbf{FaceJev}, an MLLM-based method for face forgery detection through direct visual decisions. Our design builds on Jev~\cite{typesafe2026jev} and Visual Jev~\cite{yu2026visualjev}, in which independent questions share a context and answer probabilities are read from the native answer logits. For facial analysis, we use a shared image encoding and two answer options per question, allowing the model to assign an anomaly probability to each part. These probabilities are then combined through a fixed rule to predict whether the input face image is real or forged.

FaceJev asks eight independent questions, one each about the eyes, eyebrows, nose, mouth, skin, hair, ears, and facial regions. Every question observes the full image and returns an anomaly probability for the named part. The detection score is the largest of the eight probabilities, so a face is classified as forged if any anomaly probability reaches 0.5 (\cref{fig:teaser}). This is equivalent to applying OR to the part decisions obtained at the same threshold. The image prediction is thus determined by the part decisions, which are learned through direct supervision and refined through policy optimization using both detection and part rewards.

Learning these decisions requires a target for every question. Image labels supervise only the final prediction, while judgments about the visual plausibility of individual facial parts remain implicit in existing textual explanations. We therefore construct \textbf{PartForensics}, a dataset with explicit binary anomaly targets for eight facial parts. We use existing textual explanations as references when annotating each facial part in the images. PartForensics contains approximately 160k real and forged images covering nine manipulation methods, with 1.27 million part targets aligned with the FaceJev questions. These targets directly supervise the part decisions, define the part reward, and support evaluation of the predicted anomalies across all eight facial parts.

After supervised fine-tuning (SFT), we use policy optimization with detection and part rewards to jointly optimize image classification and part decisions. Applying standard GRPO~\cite{shao2024deepseekmath} directly would assign each sampled output a single advantage based on its reward relative to the other outputs in the group. This advantage would be shared across the entire output, without distinguishing the contributions of its individual decisions. To provide feedback for each part decision, we propose \textbf{Forensic State-aware GRPO (FS-GRPO)}, which optimizes each part decision through an exact reward comparison with its alternative, using both detection and part rewards. For each sampled vector, we flip one part decision while keeping the other seven fixed and compare the resulting reward with the original reward to obtain its individual advantage. Because the rewards depend only on the binary decisions and their targets, both choices are evaluated exactly without another forward pass.

In conclusion, the main contributions of this paper are:
\begin{itemize}
 \item \textbf{FaceJev}, an MLLM-based method that detects face forgeries through direct visual decisions on eight facial parts, using independent questions on a shared image encoding and anomaly probabilities from the native answer logits.
 \item \textbf{PartForensics}, a dataset of 159,144 images covering nine manipulation methods with binary anomaly targets for the eight facial parts, supporting supervision, rewards, and evaluation of part decisions in face forgery detection.
 \item \textbf{FS-GRPO}, a policy optimization method that optimizes each part decision through an exact reward comparison with its alternative, using the detection and part rewards.
\end{itemize}

%% file: figures/method_float.tex
\begin{figure*}[!t]
 \centering
 \includegraphics[width=0.97\textwidth]{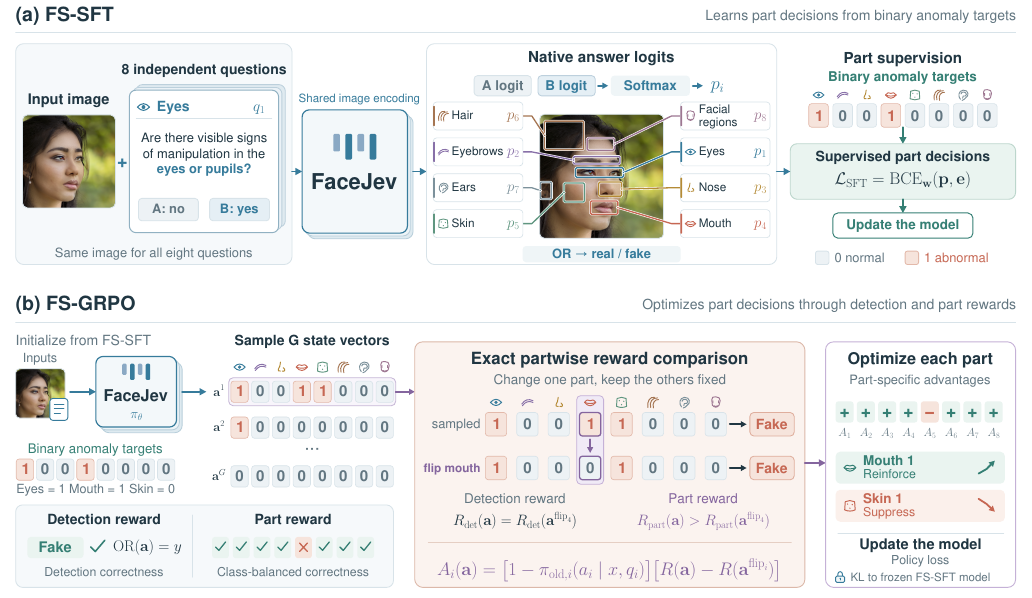}
 \caption{\textbf{Training FaceJev.} (a) FS-SFT learns anomaly probabilities from the native answer logits using the binary anomaly targets in PartForensics. (b) FS-GRPO samples binary state vectors and optimizes each part decision through an exact reward comparison with its alternative, using both detection and part rewards. Flipping the mouth preserves the forged image prediction but lowers the part reward, reinforcing the mouth decision, while the incorrect skin anomaly is suppressed. Detection uses only the image and fixed questions.}
 \label{fig:method}
\end{figure*}

%% file: sec/2_related.tex
\section{Related Work}
\label{sec:related}
\paragraph{Generalizable face forgery detection.}
Face forgery detectors seek visual cues that generalize across manipulation methods. Face X-ray detects blending boundaries~\cite{Li_2020_CVPR2}, F3-Net mines frequency-aware forgery clues~\cite{qian2020thinking}, SBI synthesizes training examples through self-blending~\cite{Shiohara_2022_CVPR}, and LAA-Net attends to localized artifacts~\cite{Nguyen_2024_CVPR}. More recent methods learn or adapt pretrained visual representations. Forensics Adapter adapts CLIP~\cite{radford2021learning} for face forgery detection~\cite{cui2024forensics}. UDD suppresses biased cues through token shuffling and mixing~\cite{udd}, FSFM learns facial representations through self-supervised pretraining~\cite{wang2025fsfm}, and Effort constrains adaptation to an orthogonal subspace to preserve pretrained knowledge~\cite{yan2024orthogonal}. GenD targets transfer across benchmarks~\cite{yermakov2025gend}, while DiCoME combines multiple views through evidential learning~\cite{kang2026dicome}. These methods learn visual features for image classification. FaceJev instead learns explicit part decisions through independent questions.

\paragraph{MLLM-based face forgery detection.}
MLLMs can recognize facial parts and assess their visual plausibility. M2F2-Det uses a visual detector to guide textual explanation generation~\cite{guo2025rethinking}, while FakeVLM learns image predictions and artifact explanations~\cite{wen2025fakevlm}. VLF-FFD fuses visual features from a forgery detector with textual representations from an MLLM through bidirectional cross-attention~\cite{peng2025mllm}. VERITAS optimizes pattern-aware reasoning with reinforcement learning~\cite{tan2025veritas}, VIGIL grounds structured reasoning in facial parts and rewards its intermediate evidence~\cite{li2026vigil}, and VRAG-DFD retrieves verifiable references to support detection~\cite{han2026vrag}. These methods typically express facial judgments through text, with supervision covering both the judgments and their wording. Individual part decisions are seldom treated as explicit learning targets. Jev~\cite{typesafe2026jev} and Visual Jev~\cite{yu2026visualjev} provide a direct decision formulation in which independent questions share a context and probabilities for a constrained set of answers are read from the native answer logits. FaceJev applies this formulation to eight facial parts and utilizes OR for the final prediction.

\paragraph{Forensic supervision.}
Benchmarks such as FaceForensics++~\cite{Rossler_2019_ICCV} and Celeb-DF~\cite{Li_2020_CVPR} provide image or video labels, and AI-Face scales labeled generated faces to a million images~\cite{lin2025aiface}. Richer annotations have followed with the rise of MLLM-based methods. DD-VQA poses common-sense questions about deepfakes~\cite{zhang2024common}, FFTG supplies structured descriptions of manipulated regions~\cite{Sun_2025_CVPR}, HydraFake provides textual explanations across diverse manipulation methods~\cite{tan2025veritas}, and MMTT pairs manipulated faces with attribution reports~\cite{lian2026mmtt}. However, judgments about the visual plausibility of individual facial parts remain embedded in textual explanations. PartForensics uses FFTG descriptions and MMTT attribution reports to guide annotation of the eight facial parts, providing binary anomaly targets for supervision, the part reward, and evaluation of part decisions.

%% file: sec/3_method.tex
\section{FaceJev}
\label{sec:method}
FaceJev detects face forgeries through direct visual decisions about eight facial parts. \Cref{sec:inference} defines the part decisions and the detection rule. Forensic State Supervised Fine-Tuning (FS-SFT) learns these decisions from binary anomaly targets (\cref{sec:sft}). Forensic State-aware Group Relative Policy Optimization (FS-GRPO) then optimizes these part decisions through detection and part rewards (\cref{sec:grpo}).

\subsection{Detection through Direct Visual Decisions}
\label{sec:inference}
Given a face image $x$, the goal is to predict its label $y\in\{0,1\}$, where 0 denotes a real image and 1 denotes a forgery. Following the Jev formulation~\cite{typesafe2026jev, yu2026visualjev}, FaceJev answers independent questions about facial parts using a shared image encoding. We read anomaly probabilities from the native answer logits, without generating any textual explanations.

We use $K=8$ fixed questions $q_1,\ldots,q_K$ for the eyes, eyebrows, nose, mouth, skin, hair, ears, and facial regions. The last category covers the forehead, cheeks, chin, and jaw. Each question uses A for no and B for yes. The eye question asks whether the eyes or pupils show visible signs of manipulation. Every question observes the full image, and no answer is used to predict another. Evidence for different parts can overlap, so a texture defect on the nose bridge may lead to abnormal predictions for both the nose and the skin.

Let $z_i^A$ and $z_i^B$ be the native answer logits for question $q_i$. Applying Softmax gives the anomaly probability $p_i(x)$:
\begin{equation}
p_i(x)=e^{z_i^B}/\big(e^{z_i^A}+e^{z_i^B}\big).
\label{eq:prob}
\end{equation}
The image detection score $s(x)$ is the largest of these probabilities. Thresholding at 0.5 returns the predicted label $\hat y$:
\begin{equation}
s(x)=\max_{i=1,\ldots,K}p_i(x),\qquad
\hat y=\ind[s(x)\geq 0.5],
\label{eq:detect}
\end{equation}
where $\ind[\cdot]$ denotes the indicator function. This is equivalent to thresholding each $p_i(x)$ at 0.5 and applying OR to the resulting part decisions. An image is therefore classified as forged if any part is predicted abnormal. We use this rule because a manipulation may affect some facial parts while leaving others intact. It directly links the detection result to part decisions learned through both FS-SFT and FS-GRPO.

\subsection{Learning the Decisions with FS-SFT}
\label{sec:sft}
In FS-SFT, shown in \cref{fig:method}(a), we supervise part decisions using binary anomaly targets in PartForensics (\cref{sec:data}). Each training image has a target vector $\mathbf e=(e_1,\ldots,e_K)$, where $e_i=1$ marks an annotated abnormality in part $i$. We minimize the weighted Binary Cross Entropy (BCE) loss:

\begingroup\footnotesize
\begin{equation}
\mathcal L_{\mathrm{SFT}}=\operatorname{BCE}_{\mathbf w}(\mathbf p,\mathbf e)=-\frac1K\sum_{i=1}^{K}\big[w_i e_i\log p_i+(1-e_i)\log(1-p_i)\big],
\label{eq:sft}
\end{equation}
\endgroup
\noindent where $p_i=p_i(x)$, $\mathbf p=(p_1,\ldots,p_K)$ contains the anomaly probabilities, and $\mathbf w=(w_1,\ldots,w_K)$ contains the weights for abnormal targets. Abnormal targets are less common than normal targets, and their frequency varies across parts. For example, the mouth is annotated as abnormal in 23,381 training images, compared with 3,588 for facial regions. We therefore set the positive weight to $w_i=\sqrt{N_{i,0}/N_{i,1}}$, where $N_{i,0}$ and $N_{i,1}$ are the numbers of training images with normal and abnormal targets for part $i$, respectively. The resulting weights are 1.36 for the mouth and 4.20 for facial regions, giving more weight to rarer abnormal targets.

\input{figures/dataset_float}

\subsection{Optimizing the Decisions with FS-GRPO}
\label{sec:grpo}
After FS-SFT, we use policy optimization with detection and part rewards to jointly optimize image classification and part decisions. Applying standard GRPO~\cite{shao2024deepseekmath} directly in this setting would assign each sampled vector one advantage based on its total reward relative to the other vectors for the same image. All decisions in that vector would share the advantage, even when some match their targets and others do not. FS-GRPO optimizes each part decision within each sampled vector through an exact reward comparison with its alternative, using both detection and part rewards.

\paragraph{State sampling and rewards.}
We sample each part decision independently as $a_i\sim\operatorname{Bernoulli}(p_i)$, where $a_i=0$ denotes normal and $a_i=1$ denotes abnormal. The probabilities of these two decisions are $1-p_i$ and $p_i$, respectively. The sampled decisions form a state vector $\mathbf a=(a_1,\ldots,a_K)\in\{0,1\}^K$, which serves as the structured action. For each training image $x$, we draw $G=8$ such vectors independently, allowing repeats. For $g=1,\ldots,G$, $\mathbf a^{g}$ denotes a sampled vector and $a_i^{g}$ its decision for part $i$. We denote the distribution for part $i$ before the current update by $\pi_{\mathrm{old},i}$, with $\pi_{\mathrm{old},i}(1\mid x,q_i)=p_i$. Each vector receives a total reward $R(\mathbf a)$ that combines detection and part rewards:

\begin{equation}
\begin{gathered}
R(\mathbf a)=R_{\mathrm{det}}(\mathbf a)+R_{\mathrm{part}}(\mathbf a),\\
R_{\mathrm{det}}(\mathbf a)=\ind[\OR(\mathbf a)=y],\\
R_{\mathrm{part}}(\mathbf a)=\frac1K\sum_{i=1}^{K}c_{i,e_i}\,\ind[a_i=e_i],
\end{gathered}
\label{eq:reward}
\end{equation}
where $c_{i,e_i}=N/(2N_{i,e_i})$, $N$ is the training set size, and $N_{i,e_i}$ counts those with target $e_i$ for part $i$. The detection reward $R_{\mathrm{det}}(\mathbf a)$ evaluates the OR prediction against the image label $y$. The part reward $R_{\mathrm{part}}(\mathbf a)$ measures agreement with binary anomaly targets, using weights to balance normal and abnormal target contributions for each facial part.

\paragraph{Exact partwise reward comparison.}
For each sampled vector $\mathbf a$, the part-specific advantage $A_i(\mathbf a)$ measures how decision $a_i$ affects the total reward. With the other decisions fixed, the baseline averages the rewards for $a_i=0$ and $a_i=1$ with sampling probabilities of $1-p_i$ and $p_i$, respectively:

\begin{equation}
A_i(\mathbf a)=R(\mathbf a)-\sum_{b\in\{0,1\}}
\pi_{\mathrm{old},i}(b\mid x,q_i)\,R(\mathbf a_{-i},b),
\label{eq:advantage}
\end{equation}
where $\mathbf a_{-i}$ contains all decisions except $a_i$, and $R(\mathbf a_{-i},b)$ is the reward after setting $a_i=b$ for $b\in\{0,1\}$. Evaluating both possible values gives an exact baseline. Let $\mathbf a^{\mathrm{flip}_i}$ denote $\mathbf a$ with $a_i$ replaced by $1-a_i$. \cref{eq:advantage} simplifies to
\begin{equation}
A_i(\mathbf a)=\big[1-\pi_{\mathrm{old},i}(a_i\mid x,q_i)\big]
\big[R(\mathbf a)-R(\mathbf a^{\mathrm{flip}_i})\big].
\label{eq:binaryadv}
\end{equation}
For the flipped state vector, we apply OR again and evaluate both rewards using the image label and part targets. These calculations use the binary decisions and existing targets, so no additional forward pass is needed. For a fixed reward difference, the factor $1-\pi_{\mathrm{old},i}(a_i\mid x,q_i)$ decreases the advantage magnitude for more probable sampled decisions.

In \cref{fig:method}(b), the eye and mouth targets are 1 and the skin target is 0, while the sampled vector marks all three as abnormal. Flipping either the mouth or the skin decision leaves the OR image prediction unchanged. Flipping the mouth decision lowers the part reward, giving a positive advantage, while flipping the skin decision raises it, giving a negative advantage. The two decisions thus receive opposite optimization signals despite the same image prediction.

\paragraph{Policy update.}
With sampled decisions and advantages fixed, we optimize the shared model using the policy loss:

\begin{equation}
\mathcal L_{\mathrm{GRPO}}=-\frac1{GK}\sum_{g=1}^{G}\sum_{i=1}^{K}
A_i(\mathbf a^{g})\log\pi_{\theta,i}\big(a_i^{g}\mid x,q_i\big),
\label{eq:policy}
\end{equation}
where $\theta$ denotes the trainable model parameters and $\pi_{\theta,i}$ is the corresponding binary distribution for part $i$. Each sampled group is used for a single update, without standardizing the advantages within each group or clipping probability ratios. We also use KL regularization toward the frozen FS-SFT model to stabilize training. The full objective and training procedure are given in the supplementary material.

%% file: figures/dataset_float.tex
\begin{figure*}[!t]
 \centering
 \includegraphics[width=0.97\textwidth]{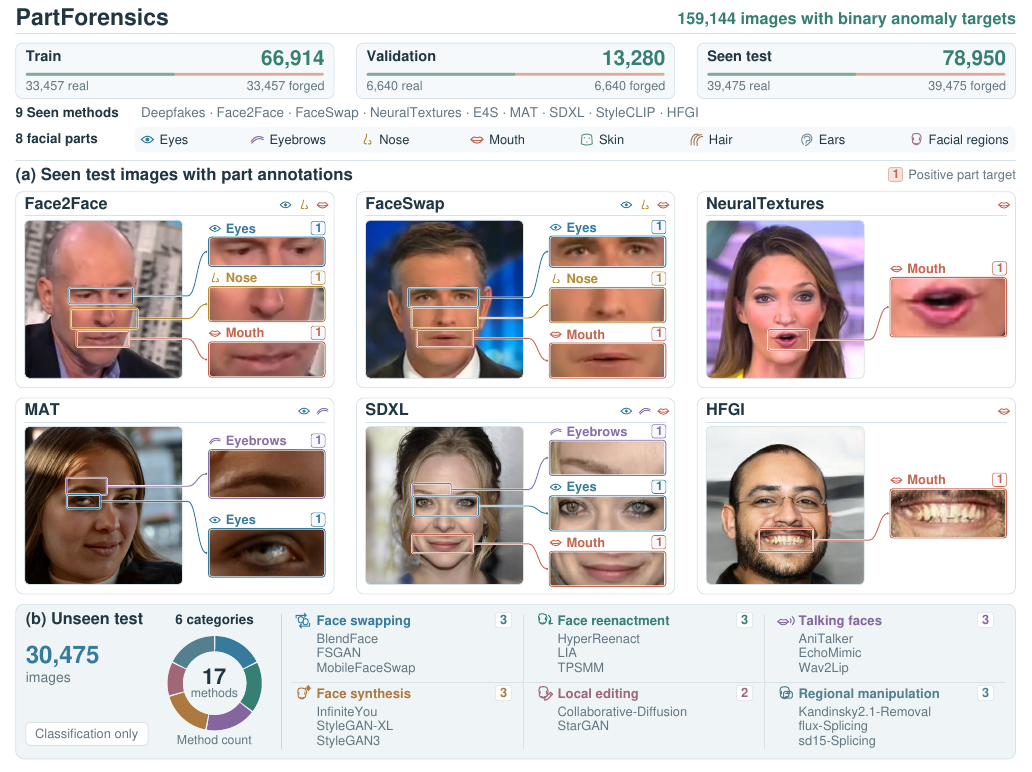}
 \caption{\textbf{PartForensics.} Top: 159,144 images with binary anomaly targets for eight facial parts, in three splits balanced between real and forged faces. (a) Examples of forged faces from the Seen test set. Boxes and crops are included only to visualize parts with positive targets. (b) The separate Unseen test set covers 17 methods in six categories, and the ring shows the number of methods in each of the categories.}
 \label{fig:dataset}
\end{figure*}

%% file: sec/4_dataset.tex
\section{PartForensics}
\label{sec:data}
PartForensics provides binary anomaly targets for the eight facial parts to supervise the direct visual decisions in FaceJev. Each sample consists of a face image $x$, its real/fake label $y$, and eight binary anomaly targets $\mathbf e$ aligned with the independent part questions. The targets supervise the part decisions in FS-SFT, define the part reward in FS-GRPO, and enable evaluation of part decisions on Seen test images.

\paragraph{Construction.}
We obtain binary anomaly targets for the eight facial parts by inspecting images with existing textual explanations as references. We use FaceForensics++~\cite{Rossler_2019_ICCV} images with structured FFTG~\cite{Sun_2025_CVPR} descriptions and MMTT~\cite{lian2026mmtt} images with attribution reports. We select these sources because the annotation procedures of FFTG and MMTT are well suited to identifying forgery artifacts in facial parts, making their explanations useful references for part annotation. MMTT contains only forged images, so we supplement this component with real faces from CelebAMask-HQ~\cite{lee2020celebamask} and FFHQ~\cite{karras2019stylegan}. For forged images from both sources, DeepSeek-V4.1-Flash~\cite{deepseek2026v41flash} inspects each image together with its explanation. It checks whether the described abnormalities are visible in the corresponding facial parts and also examines parts not mentioned in the text. The model assigns a binary anomaly target to each of the eight parts, with 0 for normal and 1 for abnormal. Real images receive all-zero part targets. At detection time, FaceJev receives only the image and fixed questions. Further annotation details are provided in the supplementary material.

\input{tables/main_results}

\paragraph{Composition.}
PartForensics contains \textbf{159,144 images} with 1,273,152 binary anomaly targets. The Train, Validation, and Seen test splits hold 66,914, 13,280, and 78,950 images, each with equal numbers of real and forged faces. The nine manipulation methods are Deepfakes, Face2Face~\cite{thies2016face2face}, FaceSwap, NeuralTextures~\cite{thies2019deferred}, E4S~\cite{liu2023fine}, MAT~\cite{li2022mat}, SDXL~\cite{podell2024sdxl}, StyleCLIP~\cite{patashnik2021styleclip}, and HFGI~\cite{wang2022high}. They cover video face manipulation, identity swapping, attribute and expression editing, and inpainting. The examples in \cref{fig:dataset}(a) show blurred lip boundaries in NeuralTextures, distorted eye and eyebrow details in MAT, and fused tooth details in HFGI. Detailed counts by source, method, and facial part are provided in the supplementary material.

\paragraph{Unseen evaluation.}
To evaluate generalization, we use a separate Unseen test set of 30,475 images with image labels only (\cref{fig:dataset}(b)). It includes 17 manipulation methods absent from training across six categories: face swapping, face reenactment, talking faces, face synthesis, local editing, and regional manipulation. The collection covers diverse manipulation categories and methods developed at different times, providing a broad test of generalization to unseen face manipulations. The supplementary material provides the complete list of manipulation methods and their image counts.

%% file: tables/main_results.tex
\begin{table*}[t]
\centering
\caption{\textbf{Face forgery detection in BACC (\%).} Seen Mean weights the nine manipulation methods equally, and Unseen Mean weights the six categories equally. Reenact. denotes face reenactment. Avg is the mean of Seen Mean and Unseen Mean. \best{Bold} and \second{underlined} values denote the best and second best results, respectively, in each column. FaceJev achieves the highest Mean BACC on both Seen and Unseen.}
\label{tab:main}
\tablefont
\setlength{\tabcolsep}{3pt}
\begin{tabularx}{\textwidth}{l*{9}{Y}}
\toprule
 & Seen & \multicolumn{7}{c}{Unseen} & \\
\cmidrule{2-2}\cmidrule(l{6pt}r{6pt}){3-9}
Method & Mean & Swapping & Reenact. & Talking & Synthesis & Editing & Regional & Mean & Avg \\
\midrule
\grouprow{10}{No training on this task}
Qwen3.5-4B~\cite{qwen2026qwen35} & 51.50 & 51.34 & 33.10 & 49.81 & 52.84 & 59.70 & 49.18 & 49.33 & 50.41 \\
Visual Jev~\cite{yu2026visualjev} & 53.89 & 52.31 & 50.02 & 50.02 & 56.37 & 55.51 & 52.75 & 52.83 & 53.36 \\
DeepSeek-V4.1-Flash~\cite{deepseek2026v41flash} & 55.65 & 52.41 & 52.27 & 51.38 & 60.34 & 57.14 & 52.88 & 54.40 & 55.03 \\
\midrule
\grouprow{10}{Visual detectors}
UDD~\cite{udd} & 97.00 & 75.08 & 71.54 & 64.54 & 77.24 & 50.57 & 82.83 & 70.30 & 83.65 \\
FSFM~\cite{wang2025fsfm} & 89.23 & \second{83.47} & 70.71 & 61.86 & 78.84 & 60.91 & 51.49 & 67.88 & 78.55 \\
Effort~\cite{yan2024orthogonal} & 95.51 & 80.70 & 74.51 & 64.99 & 82.35 & 81.18 & 68.46 & 75.37 & 85.44 \\
GenD~\cite{yermakov2025gend} & 90.31 & 83.32 & \second{75.70} & 70.02 & 78.58 & \best{87.62} & 53.81 & 74.84 & 82.57 \\
DiCoME~\cite{kang2026dicome} & 95.16 & \best{86.01} & 74.82 & \second{72.63} & 80.66 & \second{85.98} & 61.76 & \second{76.98} & \second{86.07} \\
\midrule
\grouprow{10}{MLLM-based methods}
FakeVLM~\cite{wen2025fakevlm} & \second{97.39} & 75.23 & 63.79 & 61.91 & \second{85.78} & 61.68 & \second{96.17} & 74.09 & 85.74 \\
VERITAS~\cite{tan2025veritas} & 95.06 & 71.39 & 68.81 & 66.00 & 80.55 & 57.17 & 95.08 & 73.17 & 84.12 \\
\midrule
\rowcolor{palerow}
\textbf{FaceJev} (ours) & \best{97.53} & 80.12 & \best{82.12} & \best{80.85} & \best{86.02} & 76.88 & \best{99.40} & \best{84.23} & \best{90.88} \\
\bottomrule
\end{tabularx}
\end{table*}

%% file: sec/5_experiments.tex
\section{Experiments}
\label{sec:experiments}
We first describe the experimental setup (\cref{sec:setup}) and compare detection performance on Seen and Unseen manipulations (\cref{sec:results}). Ablation studies examine the contributions of FS-SFT and FS-GRPO (\cref{sec:ablation}), followed by a comparison of part supervision across backbones (\cref{sec:backbone}). We also assess part prediction quality using quantitative metrics and visual examples (\cref{sec:states}) and compare inference cost alongside detection performance across methods (\cref{sec:inference_cost}).

\subsection{Experimental Setup}
\label{sec:setup}
\paragraph{Protocol and metrics.}
All trained models use the 66,914 images of the PartForensics training split. Seen evaluation uses the 78,950 Seen test images of the nine training methods, and Unseen evaluation uses the 30,475 images of the 17 unseen manipulation methods described in \cref{sec:data}. We report balanced accuracy (BACC), the mean of the true positive and true negative rates, at a fixed threshold of 0.5. Seen Mean averages the nine methods equally. For Unseen, we average source subsets within each method and methods within each category, and Unseen Mean weights the six categories equally. Avg is the mean of Seen Mean and Unseen Mean. We also report Part ACC and Part AUC for FaceJev against the binary anomaly targets of the Seen test images.

\input{tables/ablation}

\paragraph{Compared methods.}
We compare FaceJev with five visual detectors, UDD~\cite{udd}, FSFM~\cite{wang2025fsfm}, Effort~\cite{yan2024orthogonal}, GenD~\cite{yermakov2025gend}, and DiCoME~\cite{kang2026dicome}, and two MLLM-based methods, FakeVLM~\cite{wen2025fakevlm} and VERITAS~\cite{tan2025veritas}. These methods are trained on our training split with their default settings and evaluated on the same test images. We also evaluate three models without training on this task, including Qwen3.5-4B~\cite{qwen2026qwen35}, Visual Jev~\cite{yu2026visualjev}, and DeepSeek-V4.1-Flash~\cite{deepseek2026v41flash}.

\paragraph{Implementation.}
FaceJev adapts the vision and language components of Qwen3.5-4B~\cite{qwen2026qwen35} with LoRA~\cite{hu2022lora} of rank 128 and scaling 256, at a 512$\times$512 pixel budget, and keeps the vision-language merger frozen. Training uses AdamW~\cite{loshchilov2017decoupled}, BF16, and a global batch of 128 images on eight A100 80GB GPUs. FS-SFT trains for two epochs with a learning rate of $5\times10^{-5}$. FS-GRPO then trains for one epoch with a fresh optimizer, learning rates of $10^{-6}$ and $10^{-5}$ for the vision and language components, respectively, and $G=8$ sampled state vectors per image. Additional training details are provided in the supplementary material.

\subsection{Main Results}
\label{sec:results}
\Cref{tab:main} shows that FaceJev achieves the highest Seen Mean BACC of 97.53\% and an Unseen Mean BACC of 84.23\%, outperforming the strongest baseline on Unseen methods by 7.25 percentage points. Although the MLLM-based methods FakeVLM and VERITAS perform well on Seen manipulations, their Unseen scores fall to 74.09\% and 73.17\%, respectively, well below FaceJev. These comparisons demonstrate the effectiveness of learning explicit part decisions for generalizable face forgery detection. Notably, models without task training reach 49.33\%--54.40\% Unseen Mean BACC, with Visual Jev at 52.83\% using the same eight questions. The interface alone does not solve the task, and the following ablations examine how to learn the part decisions.

\subsection{Ablation Study}
\label{sec:ablation}
\paragraph{Learning the decisions.}
\Cref{tab:ablation}(a) compares five ways of supervising Qwen3.5-4B. Binary SFT trains the model to answer real or fake and reaches 78.57\%. Adding textual explanations to image-label supervision gives 78.02\%. FS-SFT directly supervises the eight part decisions with binary anomaly targets and reaches 80.84\%, 2.27 and 2.82 points above these alternatives, without supervising an image-level answer. Both design choices of FS-SFT contribute. Packing all eight questions into one sequence lowers Unseen Mean by 3.12 points, supporting the use of independent questions on a shared image encoding. Removing the positive weighting lowers Unseen Mean by 2.40 points, supporting the use of class weighting to account for the imbalance between normal and abnormal targets for each of the eight facial parts.

\paragraph{Optimizing the decisions.}
\Cref{tab:ablation}(b) compares policy optimization variants. FS-GRPO raises Unseen Mean from 80.84\% to 84.23\% and improves every category, by 1.47 to 4.34 points. Standard GRPO, which shares one advantage across a sample, reaches 82.97\%. FS-GRPO surpasses it by 1.26 points on average and in all six categories, supporting the optimization of individual part decisions through exact partwise reward comparisons. Both detection and part rewards contribute. Without the part reward, optimization relies on detection rewards alone, and Unseen Mean falls to 83.29\%. Without the detection reward, optimization uses only agreement with part targets, and it falls to 82.29\%. The full method outperforms both variants in every category. Running FS-GRPO from the base model reaches only 63.75\%, supporting the use of FS-SFT to initialize the part decisions before policy optimization with the two rewards.

\input{tables/backbone}

\subsection{Part Supervision across Backbones}
\label{sec:backbone}
To compare the effects of part supervision across backbones, we train two visual encoders, ConvNeXt-B~\cite{Liu_2022_CVPR} and ViT-B/16~\cite{dosovitskiy2021vit}, initialized from ImageNet~\cite{deng2009imagenet} pretraining and fine-tuned in full at 512$\times$512 resolution. For each backbone, the Binary and Parts variants use the same training settings. The Binary model predicts the image label with a single logit, while the Parts model predicts the eight PartForensics targets with eight logits from one pooled feature. The Parts model uses the weights in \cref{eq:sft} for supervision and the max/OR rule in \cref{eq:detect} for image classification. This comparison follows the same distinction between Binary SFT and FS-SFT on the Qwen3.5-4B backbone used in FaceJev.

\Cref{tab:backbone} shows that the effect of part supervision depends on the backbone. It lowers Unseen BACC by 3.08 and 3.30 percentage points on ConvNeXt-B and ViT-B/16, respectively, but improves it by 2.27 points on Qwen3.5-4B. The visual encoders predict all eight targets from one pooled feature, whereas FaceJev answers independent questions about the named facial parts using a shared image encoding. These results support using explicit part questions to adapt the facial knowledge of an MLLM for forgery detection. Applying FS-GRPO after FS-SFT yields a further gain of 3.39 percentage points, supporting policy optimization for refining the part decisions learned through supervision.

\subsection{Quality of the Part Decisions}
\label{sec:states}
\input{tables/state_analysis}
We evaluate the eight part decisions against the binary anomaly targets in PartForensics (\cref{tab:states}). FaceJev reaches 89.04\% Part ACC and 94.94\% Part AUC, showing that most part decisions agree with the annotated targets and that the predicted probabilities distinguish normal from abnormal parts effectively. Part AUC exceeds 89\% for every Seen manipulation method, supporting the effectiveness of the part predictions across all nine methods. Overall, part prediction performance is stronger on the four FaceForensics++ methods, Deepfakes, Face2Face, FaceSwap, and NeuralTextures, than on the MMTT methods, E4S, MAT, SDXL, StyleCLIP, and HFGI, in terms of the Part ACC and Part AUC metrics.

\input{figures/qualitative_float}
\Cref{fig:seen_part_predictions} shows FaceJev's part decisions on eight Seen test images. The displayed predictions include mouth abnormalities and combinations involving the eyes, nose, hair, skin, and eyebrows. Each displayed probability exceeds the detection threshold of 0.5, making explicit which part decisions support a forged image prediction under the OR rule.

\subsection{Inference Cost}
\label{sec:inference_cost}
\input{tables/efficiency}
\Cref{tab:efficiency} compares FaceJev with FakeVLM and VERITAS, followed by the Binary SFT and Explanation SFT ablations. FakeVLM and VERITAS produce both a binary score and a textual explanation, processing 2.93 and 2.39 images per second, respectively. Among the SFT variants, Binary SFT is fastest at 14.60 images per second because it returns only a binary score, while Explanation SFT processes 1.28 images per second with explanation generation included. With a shared image encoding, FaceJev predicts anomaly probabilities for all eight facial parts and uses the maximum as the image detection score. It processes 8.71 images per second without generating textual explanations. It is faster than every compared method except Binary SFT and achieves the highest Unseen BACC of 84.23\%, offering a favorable balance between its detection performance and inference cost.

%% file: tables/ablation.tex
\begin{table*}[t]
\centering
\caption{\textbf{Ablations on Unseen methods in BACC (\%).} All configurations use Qwen3.5-4B and the same PartForensics training images. $\Delta$ is the change of Unseen Mean relative to the last row of each panel, and \textbf{bold} marks the best value for every metric within each panel. FS-SFT achieves the highest Unseen Mean among supervision variants in panel (a), while FS-GRPO leads all six categories in panel (b).}
\label{tab:ablation}
\tablefont
\setlength{\tabcolsep}{2pt}
\begin{tabularx}{\textwidth}{ll*{8}{Y}}
\toprule
Configuration & Change & Swapping & Reenact. & Talking & Synthesis & Editing & Regional & Mean & $\Delta$ \\
\midrule
\grouprow{10}{(a) Learning the decisions}
Binary SFT & image label only & 73.90 & 70.05 & 67.60 & 83.08 & \best{78.52} & \best{98.28} & 78.57 & \dn{2.27} \\
Explanation SFT & label + explanation & 75.51 & 70.87 & 68.02 & 82.65 & 76.20 & 94.88 & 78.02 & \dn{2.82} \\
Joint queries & one joint sequence & 74.80 & 69.88 & 67.73 & \best{84.19} & 73.46 & 96.24 & 77.72 & \dn{3.12} \\
No weighting & $w_i=1$ & \best{76.60} & 75.48 & 76.36 & 82.05 & 66.03 & 94.14 & 78.44 & \dn{2.40} \\
\rowcolor{palerow}
FS-SFT & independent, weighted & 75.78 & \best{78.00} & \best{76.90} & 82.14 & 75.41 & 96.82 & \best{80.84} &  \\
\midrule
\grouprow{10}{(b) Optimizing the decisions}
No FS-SFT & from base model & 65.45 & 57.18 & 49.47 & 78.23 & 60.34 & 71.81 & 63.75 & \dn{20.48} \\
FS-SFT + GRPO & shared advantage & 77.91 & 80.60 & 79.66 & 84.97 & 75.90 & 98.78 & 82.97 & \dn{1.26} \\
No part reward & $R=R_{\mathrm{det}}$ & 78.99 & 81.36 & 79.89 & 85.79 & 75.89 & 97.81 & 83.29 & \dn{0.94} \\
No detection reward & $R=R_{\mathrm{part}}$ & 77.44 & 79.58 & 78.83 & 83.83 & 75.93 & 98.15 & 82.29 & \dn{1.94} \\
\rowcolor{palerow}
FaceJev & $R=R_{\mathrm{det}}+R_{\mathrm{part}}$ & \best{80.12} & \best{82.12} & \best{80.85} & \best{86.02} & \best{76.88} & \best{99.40} & \best{84.23} &  \\
\bottomrule
\end{tabularx}
\end{table*}

%% file: tables/backbone.tex
\begin{table}[t]
\centering
\caption{\textbf{Part supervision across backbones.} Unseen BACC (\%) with image labels (Binary) or eight part targets and the OR readout (Parts). For Qwen3.5-4B, the SFT row reports Binary SFT and FS-SFT, respectively. FaceJev adds FS-GRPO to FS-SFT. $\Delta$ is Parts minus Binary, with all differences expressed in percentage points.}
\label{tab:backbone}
\tablefont
\setlength{\tabcolsep}{2pt}
\begin{tabularx}{\columnwidth}{l*{4}{Y}}
\toprule
Model / training & Params & Binary & Parts & $\Delta$ \\
\midrule
\grouprow{5}{Vision encoder with linear heads}
ConvNeXt-B~\cite{Liu_2022_CVPR} & 88M & 67.90 & 64.82 & \dn{3.08} \\
ViT-B/16~\cite{dosovitskiy2021vit} & 86M & 70.83 & 67.53 & \dn{3.30} \\
\midrule
\grouprow{5}{MLLM with direct visual decisions}
Qwen3.5-4B, SFT & 4B & 78.57 & 80.84 & \up{2.27} \\
\rowcolor{palerow}
Qwen3.5-4B, FaceJev & 4B & -- & \best{84.23} & \\
\bottomrule
\end{tabularx}
\end{table}

%% file: tables/state_analysis.tex
\begin{table}[t]
\centering
\caption{\textbf{Image and part decisions of FaceJev on Seen test (\%).} Part metrics pool all image-part pairs of a method. Mean is the unweighted average across the nine Seen manipulation methods.}
\label{tab:states}
\tablefont
\setlength{\tabcolsep}{3pt}
\begin{tabularx}{\columnwidth}{l*{3}{Y}}
\toprule
Method & Image BACC & Part ACC & Part AUC \\
\midrule
Deepfakes & 98.09 & 99.15 & 99.81 \\
Face2Face & 94.83 & 94.50 & 97.86 \\
FaceSwap & 95.37 & 96.31 & 98.91 \\
NeuralTextures & 91.20 & 94.62 & 96.66 \\
E4S & 99.82 & 77.77 & 89.73 \\
MAT & 99.05 & 81.83 & 91.24 \\
SDXL & 99.79 & 79.69 & 90.81 \\
StyleCLIP & 99.81 & 89.53 & 95.32 \\
HFGI & 99.84 & 87.95 & 94.09 \\
\midrule
\rowcolor{palerow}
\textbf{Mean} & \textbf{97.53} & \textbf{89.04} & \textbf{94.94} \\
\bottomrule
\end{tabularx}
\end{table}

%% file: figures/qualitative_float.tex
\begin{figure}[!t]
 \centering
 \includegraphics[width=\columnwidth]{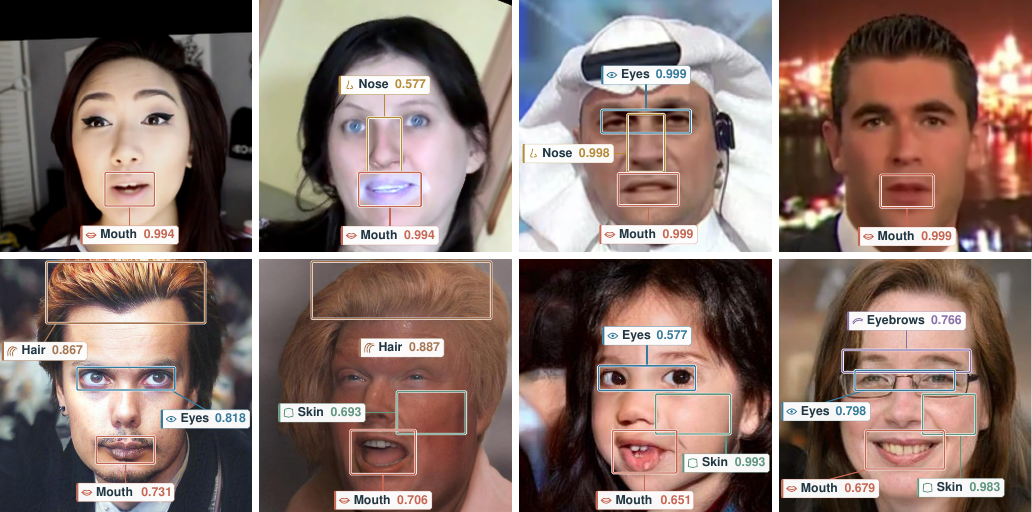}
 \caption{\textbf{Part decisions of FaceJev on eight Seen test images.} Labels show predicted anomaly probabilities, and boxes mark the corresponding facial parts in the input face images for illustration.}
 \label{fig:seen_part_predictions}
\end{figure}

%% file: tables/efficiency.tex
\begin{table}[t]
\centering
\caption{\textbf{Inference output, throughput, and Unseen BACC (\%).} We compare FakeVLM and VERITAS with the Binary SFT and Explanation SFT variants in \cref{tab:ablation}(a), and the FaceJev method. Throughput (img/s) is measured on one A100 GPU using BF16 and batch size 8. \best{Bold} highlights the best result for each metric.}
\label{tab:efficiency}
\tablefont
\setlength{\tabcolsep}{3pt}
\begin{tabularx}{\columnwidth}{l>{\raggedright\arraybackslash}Xcc}
\toprule
Method & Output & Throughput & Unseen \\
\midrule
\grouprow{4}{Compared methods}
FakeVLM~\cite{wen2025fakevlm} & Score + explanation & 2.93 & 74.09 \\
VERITAS~\cite{tan2025veritas} & Explanation + score & 2.39 & 73.17 \\
\midrule
\grouprow{4}{SFT ablations}
Binary SFT & Binary score & \best{14.60} & 78.57 \\
Explanation SFT & Score + explanation & 1.28 & 78.02 \\
\midrule
\rowcolor{palerow}
FaceJev & Part decisions + score & 8.71 & \best{84.23} \\
\bottomrule
\end{tabularx}
\end{table}

%% file: sec/6_conclusion.tex
\section{Conclusion}
\label{sec:conclusion}
We presented FaceJev, an MLLM-based method for face forgery detection through direct visual decisions. Eight independent questions about facial parts use a shared image encoding of the full input image, with anomaly probabilities read from the native answer logits. A face is classified as forged if any part is predicted to be abnormal. PartForensics provides binary anomaly targets for supervising these decisions, and FS-GRPO optimizes each decision through an exact reward comparison with its alternative, using both detection and part rewards. Ablations support these design choices. FaceJev achieves the highest Seen and Unseen Mean BACC among the compared methods, surpassing the strongest Unseen baseline by 7.25 percentage points. It also provides faster inference than the compared methods that generate textual explanations. These results demonstrate the potential of FaceJev for face forgery detection.